\pdfoutput=1
\documentclass[review]{elsarticle}

\usepackage{lineno,hyperref}
\hypersetup{pdfauthor=author}

\usepackage{etoolbox}
\makeatletter
\patchcmd\ps@pprintTitle
  {\fi\fi\fi\fi}
  {\fi\fi\fi\fi
   \afterassignment\fix@elsarticle\let\@tempa}
  {}{\FailedToPatch}
\def\fix@elsarticle{\iffalse{\fi}\romannumeral-`0}
\makeatother

\journal{Knowledge-Based Systems}

\usepackage{soul}
\usepackage{color}

\usepackage[numbers]{natbib}

\usepackage{multirow}
\usepackage{amssymb}
\usepackage{bbding}
\usepackage{array}
\usepackage{makecell}
\usepackage{enumitem}

\begin{document}

\begin{frontmatter}

\title{Knowledge-aware Document Summarization: A Survey of Knowledge, Embedding Methods and Architectures}





\author[adelaideuni]{Yutong Qu\corref{mycorrespondingauthor}}
\cortext[mycorrespondingauthor]{Corresponding author}
\ead{yutong.qu@adelaide.edu.au}

\author[adelaideuni]{Wei Emma Zhang\corref{mycorrespondingauthor}}
\ead{wei.e.zhang@adelaide.edu.au}

\author[macquarieuni]{Jian Yang}
\author[jdcom]{Lingfei Wu}
\author[macquarieuni]{Jia Wu}

\address[adelaideuni]{School of Computer Science, The University of Adelaide, SA 5005, Australia}
\address[macquarieuni]{School of Computing, Macquarie University, NSW 2109, Australia}
\address[jdcom]{JD.COM Silicon Valley Research Center, CA 94043, USA}

\begin{abstract}
Knowledge-aware methods have boosted a range of natural language processing applications over the last decades. With the gathered momentum, knowledge recently has been pumped into enormous attention in document summarization, one of natural language processing applications. Previous works reported that knowledge-embedded document summarizers excel at generating superior digests, especially in terms of informativeness, coherence, and fact consistency. This paper pursues to present the first systematic survey for the state-of-the-art methodologies that embed knowledge into document summarizers. Particularly, we propose novel taxonomies to recapitulate knowledge and knowledge embeddings under the document summarization view. We further explore how embeddings are generated in embedding learning architectures of document summarization models, especially of deep learning models. At last, we discuss the challenges of this topic and future directions.

\end{abstract}


\end{frontmatter}


\section{Introduction}

With the exponential burst of textual data, demands in condensing voluminous text contents have been ubiquitous, bringing document summarization one of the most immensely researched fields in Natural Language Processing (NLP). 
Document Summarization (DS) aims to generate an abridged version of single or multiple topic-related texts as concise and coherent as possible while preserving the salient and factually consistent information \cite{ma2020MDSsur}. 
The document summarization task with a single input document is known as the Single Document Summarization (SDS). By contrast, the Multi-Document Summarization (MDS) task emphasizes synthesizing a large number of topic-related documents to generate a compressed summary from various times and perspectives.
%
In addition, there are two general methods in document summarization: 1) the Extractive Document Summarization (EDS) method respects the lexicon of the original text, regarding the summary formation is verbatim by key words and phrases selected from the source corpus; and 2) the Abstractive Document Summarization (ADS) method respects the semantics of the original text, regarding the summary construction is by rephrasing texts according to the comprehension of text substances.
%
%
Generally, a document summarization model is to achieve the following goals \cite{elkassas2021DSsur}:
%
%
\begin{itemize}
    \item \textbf{G1.} Coverage: A document summarization model aims to generate a comprehensive summary that covers all the main and noteworthy contents of the input text(s);
    \item \textbf{G2.} Non-redundancy: A document summarization model aims to generate a precise and concise summary without any redundant or meaninglessly repeated information;
    \item \textbf{G3.} Readability: A document summarization model aims to generate a smooth and logical summary composed by human-readable and coherent sentences to the viewer;
\end{itemize}
For multi-document summarization models, an additional goal is \cite{ferreira2014MDSsur}:
\begin{itemize}
    \item \textbf{G4.} Relevancy: A multi-document summarization model aims to identify related information within multiple input texts while generating the summary.
\end{itemize}
Recently, knowledge utilization in the summarization models has exhibited a huge potential for promoting the summarizer performance in terms of \textbf{G1} to \textbf{G4} and fueled one more document summarization capacity target:
\begin{itemize}
    \item \textbf{G5.} Factual Consistency: A document summarization model aims to generate a consistent summary that obeys text facts and the commonsense of the real world.
\end{itemize}
The goal \textbf{G5} also reflects that the knowledge refers to the information acquired from facts and commonsense in source corpora and external sources, which normally can be captured in knowledge graphs \cite{hogan2021kgesur}.
In contrast to novel auxiliary knowledge, such as the timeline or visual information \cite{newkninDS2021}, the knowledge, in general, focuses more on human knowledge from a linguistic perspective, adapting to a broader range of standard summarization tasks.
Throughout the knowledge usage in document summarization, from word-level knowledge \cite{han2016sekg} to document-level knowledge \cite{yasunaga2017mesg} and from internal knowledge \cite{tan2017saug} to external fact knowledge \cite{gunel2019sakg}, we observe that various-formed knowledge appears and is incorporated for document summarization in different ways.
In addition, empirical evidence from studies \cite{ji2021sakg,tang2020sekg,wu2021masg,chen2021mesdg} reported a worthwhile potentiality of leveraging different kinds of knowledge in both extractive and abstractive document summarization methods for single or multiple inputs.
Also of note are the possibility and motivated envisagement of effectively blending a variety of knowledge for document summarization to enrich the fact and commonsense consistency of generated summaries.
%
%
However, there is no existing work to summarize these research contributions. 
To fill this gap, we systematically investigate the knowledge and knowledge embedding methodologies under the document summarization view and report the results in this survey paper. 

\begin{table*} \scriptsize
\centering
\begin{tabular}{ccc}
\hline
\textbf{Surveys} & \textbf{Coverage} & \textbf{Domain} \\
\hline
\citet{wang2017kgesur} & KGE & NLP    \\

\citet{cai2018gesur} & KG; GE & AI    \\

\citet{xu2021gesur} & KG; GE & AI   \\

\citet{ji2021kgesur} & KG; KGE & NLP    \\

\citet{hogan2021kgesur} & KG; KGE & NLP    \\

\hline
Ours & K; KE & DS    \\
\hline
\end{tabular} 
\caption{Outline of comparisons between existing relative surveys and ours. \textbf{K}, \textbf{G}, and \textbf{E} denote knowledge, graph, and embedding, respectively.}
\label{tab:difference}
\end{table*}

\noindent \textbf{Comparisons to other surveys.} 
The related works to our paper are surveys on knowledge, knowledge graphs, and knowledge embeddings in artificial intelligence applications.
\citet{wang2017kgesur} summarized specifically on knowledge graph embedding with a systematic review of existing embedding techniques in a range of the natural language processing applications, such as relation extraction and question answering.
\citet{cai2018gesur} proposed a classification of graph embedding work based on problem settings with descriptions of graph embedding techniques and applications in the artificial intelligence field, such as graph classification and graph visualization.
Similarly, \citet{xu2021gesur} broadly categorized graph embedding methods according to base factors of graph embedding methods, such as matrix factorization, random walk, and neural network. Also, \citet{xu2021gesur} introduced representative real-world application examples from academia and industries in artificial intelligence.
\citet{ji2021kgesur} provided a technical overview of knowledge graphs with knowledge graph embedding and introduced downstream knowledge-aware natural language processing applications such as question answering and recommendation systems.
\citet{hogan2021kgesur} comprehensively introduced diverse concepts and aspects of knowledge graphs. Besides, \citet{hogan2021kgesur} distinguished open-source knowledge graphs as open knowledge graphs and regarded the internally constructed and utilized knowledge graphs as enterprise knowledge graphs. 
%
%
Table \ref{tab:difference} presents an intuitive comparison between these  relevant literature review articles and this paper. In a brief summary, the recent related works focus more on introducing either graph embeddings or knowledge graphs for a wide range of artificial intelligence applications in a general manner. Neither of them thoroughly targets a systematic view for one specific application.
Differently, our survey studies a complete process of leveraging the general knowledge in a promising natural language processing application, document summarization: from acquiring knowledge to embedding knowledge, followed by how embedding learning architectures generate and work with knowledge embeddings, under the document summarization view. We select, describe, and analyze the state-of-the-art document summarization works that embed general knowledge into models, and form the first systematic literature review of this kind.
%

\noindent \textbf{Contributions of this survey.}
This survey contributes to the document summarization field with the investigation of the usage of knowledge and knowledge embeddings in document summarization models.
%
%
Specifically, our \textit{\textbf{first contribution}} is a taxonomy of the general knowledge leveraged in document summarization, presented in Section \ref{sec:knowtaxo}. In this paper, we consider all of the external relevant auxiliary information and the derived linguistic information in addition to the plain textual input as \textbf{knowledge} for document summarization, which is an expansion of the general factual knowledge. 
In our taxonomy, we broadly classify the knowledge incorporated in document summarization into four main categories: \textit{native knowledge}, \textit{linguistic knowledge}, \textit{semantic knowledge}, and \textit{topical knowledge}.
The categorization is conducted lying on the layers of the knowledge from literalness to connotation, implied in hierarchies of the documentary information from word to the full text. The sub-categories are also discussed.
Knowledge embeddings refer to low-dimensional and continuous representations of the knowledge \cite{wang2017kgesur}, profiting better ways to permit various discrete-formed knowledge to be incorporated into learning models.
Due to different-level knowledge leveraged for document summarization, a wide variety of knowledge embedding methodologies have been employed in document summarization models.
Our \textit{\textbf{second contribution}} is introducing a taxonomy of the existing knowledge embedding methodologies in document summarization tasks, introduced in Section \ref{sec:knowembdtaxo} and how document summarization embedding learning architectures generate different knowledge embeddings as discussed in Section \ref{sec:kelearnarc}.
%
Finally, we provide our envision about the future directions on the existing issues and unfilled gaps, which are aligned with the five goals of the document summarization task
in Section \ref{sec:outlook}, forming our \textit{\textbf{third contribution}},
followed by the conclusion in Section \ref{sec:conclude}.
%

\begin{table*} \scriptsize
\setlength{\tabcolsep}{4.3pt}
\centering
\begin{tabular}{ccp{0.24cm}<{\centering}p{0.2cm}<{\centering}p{0.18cm}<{\centering}p{0.22cm}<{\centering}p{0.21cm}<{\centering}p{0.21cm}<{\centering}ccc}
\hline

\multicolumn{1}{c}{ \multirow{2}{*}{\textbf{Methods}} } &
\textbf{\makecell{DS}} &
\multicolumn{7}{c}{ \textbf{\makecell{Knowledge}} } &
\multicolumn{1}{c}{ \multirow{2}{*}{\textbf{KG}} } &
\textbf{\makecell{Model}}
\\

 & 
 \multicolumn{1}{c}{\textbf{Tasks}} & \textbf{NK} & \textbf{LK} & \textbf{SK} & \textbf{DK} & \textbf{CK} & \textbf{OK} & \textbf{TK} & &
\multicolumn{1}{c}{\textbf{Architectures}} \\

\hline
\multicolumn{1}{l}{\textbf{A-SDS}} \\
%
ABS+AMR \cite{takase2016sasg} & & & & $\checkmark$ & & &   $\checkmark$ & & $\checkmark$ & Encoder-Decoder  \\
%
GAM+HBS \cite{tan2017saug} & & $\checkmark$ &  $\checkmark$ & & & & & & $\checkmark$ & Encoder-Decoder \\
BERTSUM \cite{liu2019saeng}  & A/E &  $\checkmark$ & & & & & & & & BiTransformer    \\
%
BERT+RL \cite{zhang2019sang} & &  $\checkmark$ & & & & & & & & Transformer    \\
%
GraphWriter \cite{koncelkedziorsk2019sakg} & &   $\checkmark$ & & & & &   $\checkmark$ & & $\checkmark$ & G-Transformer   \\
%
PG+PreTrained \cite{anhdt2019sang} & &    $\checkmark$ & & & & & & & & Pointer Generator    \\
%
IE+MSA \cite{guan2019sakg} & & & & & &   $\checkmark$ & & & $\checkmark$ & Encoder-Decoder  \\
%
TXL+WikiKG \cite{gunel2019sakg} & &    $\checkmark$ & & & &   $\checkmark$ & & & $\checkmark$ & Transformer-XL                  \\
%
SemSUM \cite{jin2020sasg} & &    $\checkmark$ & &  $\checkmark$ & & & & & $\checkmark$ & Transformer  \\
%
GRF \cite{jihaozhe2020sakg} & &   $\checkmark$ & & & &   $\checkmark$ & & & $\checkmark$ & GPT-2     \\
%
PGN+IDF \cite{you2021sang}  & &    $\checkmark$ & & & & & & & & Encoder-Decoder    \\
%
FASUM \cite{zhu2021sakg} & &    $\checkmark$ & & & & &   $\checkmark$ & & $\checkmark$ & Seq2Seq  \\
%
SKGSUM \cite{ji2021sakg} & &    $\checkmark$ &  $\checkmark$ & &  $\checkmark$ & &   $\checkmark$ & & $\checkmark$ & Encoder-Decoder  \\
\hline
\multicolumn{1}{l}{\textbf{A-MDS}} \\
%
Seq2Seq+MTG \cite{fan2019makg} & &   &  $\checkmark$ & & & &   $\checkmark$ & & $\checkmark$ & Transformer  \\
%
PEGASUS \cite{zhang2020mang} & &    $\checkmark$ & & & & & & & & Encoder-Decoder    \\
%
ASGARD \cite{huang2020makg}  & &    $\checkmark$ & & & & &   $\checkmark$ & & $\checkmark$ & Encoder-Decoder    \\
%
BartGraphSumm \cite{pasunuru2021masg} & &    $\checkmark$ &  $\checkmark$ & & & &   $\checkmark$ & & $\checkmark$ & BART-Long    \\
%
EMSUM \cite{zhou2021masg} & &    $\checkmark$ &  $\checkmark$ &  $\checkmark$ & & & & & $\checkmark$ & Transformer  \\
BASS \cite{wu2021masg} & S/M &    $\checkmark$ & &  $\checkmark$ & & &   $\checkmark$ & & $\checkmark$ & Encoder-Decoder    \\
\hline
\multicolumn{1}{l}{\textbf{E-SDS}} \\
%
FSGM \cite{han2016sekg} & &    $\checkmark$ &  $\checkmark$ & & &   $\checkmark$ & & & $\checkmark$ & GraphRank   \\
%
RNN+LSTM \cite{zhang2017seng} & &                      $\checkmark$ & & & & & & & & Encoder-Decoder                \\
%
HIBERT \cite{zhang2019seng}  & &    $\checkmark$ & & & & & & & & Transformer     \\
%
BERT+HGM \cite{yuan2020sesg} & &    $\checkmark$ & &  $\checkmark$ & & & & & $\checkmark$ & BERT+HGM \\
%
RST+spanBERT \cite{huang2021sedg} & &    $\checkmark$ & & &  $\checkmark$ & & & & $\checkmark$ & Longformer  \\
%
Topic-GraphSum \cite{cui2020seug} & &    $\checkmark$ & & & & & & $\checkmark$ & $\checkmark$ & GNN  \\
%
DISCOBERT \cite{xu2020sedg} & &    $\checkmark$ &  $\checkmark$ & &  $\checkmark$ & & & & $\checkmark$ & Transformer     \\
%
kg-KMTR \cite{tang2020sekg} & &    $\checkmark$ & & & &   $\checkmark$ & & & $\checkmark$ & TextRank+K-means  \\
\hline
\multicolumn{1}{l}{\textbf{E-MDS}} \\
%
GRU+GCN \cite{yasunaga2017mesg} & &    $\checkmark$ &  $\checkmark$ & &  $\checkmark$ & & & & $\checkmark$ & GNN    \\
%
STDS \cite{zheng2019meng} & &                          $\checkmark$ & & & & & & $\checkmark$ & & Encoder-Decoder    \\
%
HETERSUMGRAPH \cite{wang2020mesg} & &    $\checkmark$ &  $\checkmark$ & & & & & & $\checkmark$ & GNN  \\
\hline
\end{tabular}
\caption{List of the representative \textbf{A}bstract or \textbf{E}xtractive \textbf{S}ingle- or \textbf{M}ulti-document summarization (\textbf{DS}) methods incorporated the knowledge, indicating the usage of knowledge graphs (\textbf{KG}) and the main model architectures. The described Native Knowledge (\textbf{NK}), Lexical Knowledge (\textbf{LK}),  Syntactic Knowledge (\textbf{SK}), Discourse Knowledge (\textbf{DK}), Closed Knowledge (\textbf{CK}), Open Knowledge (\textbf{OK}), and Topical Knowledge (\textbf{TK}) are presented.
%
%
}
\label{tab:dsknowledge}
\end{table*}

\section{Knowledge Taxonomy}
\label{sec:knowtaxo}

\begin{figure*}[ht]
  \centering
  \includegraphics[width=0.5\textwidth]{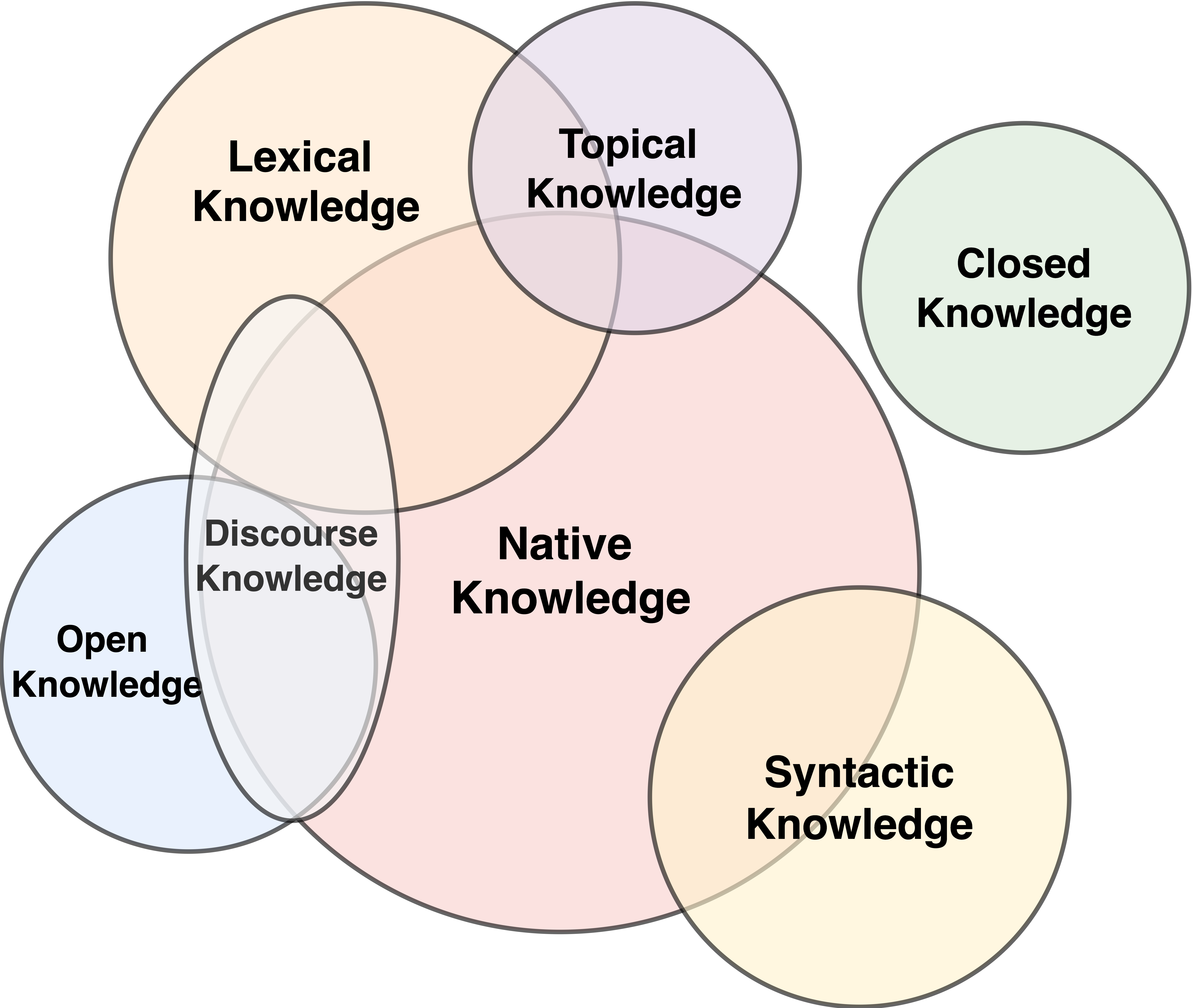}
  \caption{Knowledge categorization from the document summarization perspective. 
  %
  %
  }
  \label{fig:kg}
  \vspace{-2mm}
\end{figure*}

%
In this survey, we classify the group of knowledge incorporated in document summarization models into four main categories. Relations among each type of knowledge are illustrated in Figure \ref{fig:kg}. Moreover, the investigation of knowledge leveraged in the state-of-the-art document summarization methods is shown in Table \ref{tab:dsknowledge}, with the order of the timeline of publication. It covers the usage status of knowledge graphs, the utilized knowledge relying on our proposed knowledge classification, and the main model architectures of those reviewed methods.
%
%
The knowledge is obtained from the literal text or latent semantic space and can work alone or by merging to derive high-level information for the goals \textbf{G1} to \textbf{G5}.
\subsection{Native knowledge}
Native knowledge is the raw and plain textual data in the source text garnered without any filtration or transformation processing, such as the original words and sentences of the source text, typically leveraged as the auxiliary information \cite{zhang2020mang}. This knowledge is captured in non-graph structures, such as in the form of token vectors, directly embedded into the model. It can represent the maximal amount of the origin text information and enhance the content richness for promoting to achieve the goal \textbf{G1}.

\subsection{Linguistic knowledge}
Linguistic knowledge focuses on the source text information with an additional linguistic perspective than the native knowledge, such as the information related to the lexis, syntax, and grammar of the source text. Generally, linguistic knowledge is presented as in gauged relations among words or parsed relations of sentence dependencies.

\paragraph{Lexical knowledge} It is the estimated lexical relation knowledge among text entities (i.e., words), such as the centrality \cite{erkan2004lexrank}, textual similarity \cite{yasunaga2017mesg,fan2019makg,li2020masdg,wang2020mesg,ji2021sakg,zhou2021masg,chen2021mesdg}, semantic similarity \cite{han2016sekg,tan2017saug,zhang2020mang}, and salience \cite{yasunaga2017mesg,tan2017saug} information.
This knowledge is in the form of numerical scores, infused as weights in learning models.
It conduces to filter the relevant and salient text units for generating informative and succinct summaries, tallying with the goals \textbf{G1} and \textbf{G2}. Also, it captured word relations can enhance summary coherence for the goal \textbf{G3}.

\paragraph{Syntactic knowledge} It involves in syntactic dependency relations extracted by dependency parsers, such as the \textit{JAMR} \cite{flanigan2014jamr}, \textit{CoreNLP dependency parser} \cite{hermann2015conlppar}, and neural dependency
parser \cite{dozat2017neudpar}. 
This dependency relation forms the syntactic knowledge among words of each sentence, commonly modelled as dependency trees.
The preserved syntactic relations can assist in determining redundant units and improving the summary coherence forward to the goals \textbf{G2} and \textbf{G3}.

\paragraph{Discourse knowledge} It covers  discourse dependency relations concluded by discourse relation indicators via discovering deverbal noun references, event or entity continuations, discourse markers, or coreferent mentions \cite{yasunaga2017mesg,li2020masdg,xu2020sedg}, or gathered by discourse parsers \cite{ji2014discpar}.
This knowledge is usually formed as discourse graphs.
It contains both syntactic and semantic information, excelling at redundancy recognition and logic enhancement, profiting the goals \textbf{G2} and \textbf{G3}.

\subsection{Semantic knowledge}
Semantic knowledge concentrates on conceptual and factual information gained from the real world or extracted from the source text. It is in the form of triplet (subject, predicate, object) and typically preserved in knowledge graph (KG) or knowledge base (KB).

\paragraph{Closed knowledge} It is the lexical relationship knowledge from the existing open-source and graph-based databases that contain general commonsense and human explicit knowledge, such as \textit{WordNet} \cite{miller1995wordnet}, \textit{FrameNet} \cite{ruppenhofer2006framenet}, \textit{ConceptNet5} \cite{speerhavasi2012conceptnet}, and \textit{Wikidata} \cite{vrandei2014wikidata}. 
This triplet-formed knowledge involves 
real-world facts that are surpassed to detect inconsistent fact errors in document summarization for achieving the goal \textbf{G5}.

\paragraph{Open knowledge} It is the ever-evolving and expansible  knowledge of semantic relations extracted and accumulated from source corpora by information extraction tools, such as \textit{Open-domain Information Extraction (OpenIE)} models \cite{angeli2015lsopenie,stanovsky2018sopenie}.
Similar to close knowledge, open knowledge is also in the form of the triplet (subject, predicate, object), but with more flexible subjects, predicates and objects. 
The semantic relations held by the open knowledge could also help improve the concision and logic of the summary, promoting the goal \textbf{G5}.  

\subsection{Topical knowledge}
Topical knowledge is a latent knowledge of the source text, gained by topic models, such as the Latent Dirichlet Allocation (LDA) \cite{blei2003lda} or neural topic model (NTM) \cite{miao2017ntm}. This knowledge mainly comprises the information of the topic salience \cite{zheng2019meng} and topical relevance \cite{cui2020seug,li2020masdg}. It can indicate the phrase-level semantic information to enhance the summary coherence for the goal \textbf{G2} or imply the document-level semantic information for capturing relations among documents, benefiting the goal \textbf{G3}.

\begin{figure*}[t]
  \centering
  \includegraphics[width=1.03\textwidth]{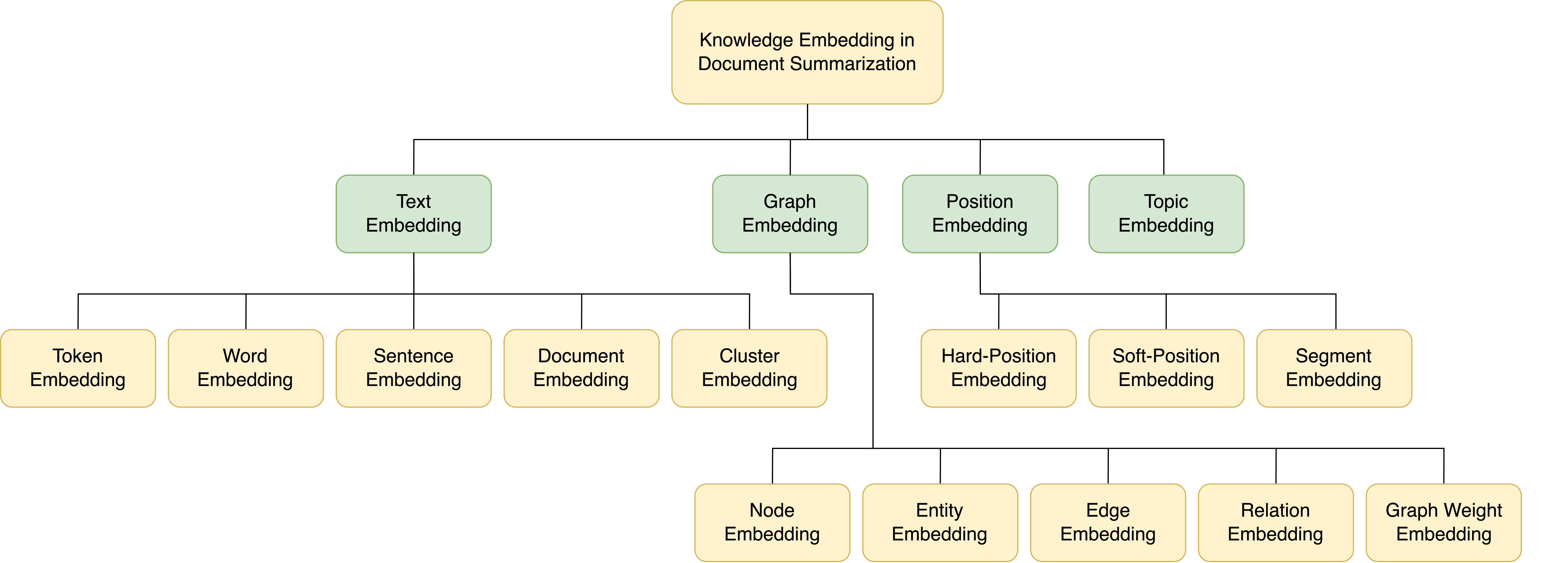}
  \caption{The taxonomy of knowledge embeddings utilized in document summarization tasks.
  %
  }
  \label{fig:embedding}
\end{figure*}

\section{Knowledge Embedding Taxonomy}
\label{sec:knowembdtaxo}
\textit{Native knowledge} is usually in the original textual form, and its embedding relies on the embedding of the textual components in the document, such as token embedding, word embedding, sentence embedding and document embedding. \textit{Linguistic knowledge} could be formed as texts or relations, and the latter is commonly modelled as a graph. Therefore, embedding linguistic knowledge covers both text embedding and graph embedding. \textit{Semantic knowledge} similarly leverages both textual embedding and graph embedding. \textit{Topical knowledge} is usually in data distribution form and requires to embed the distributions.
In order to present the knowledge embedding applied in document summarization clearly, instead of grouping the embedding methods according to knowledge categorization, 
we propose \textbf{a new taxonomy} for knowledge embedding methods, as shown in Figure \ref{fig:embedding}.

\vspace{1.5mm}
\noindent \textbf{Text Embedding} Many knowledge embedding methods in document summarization focus on using textual contents from the source corpus.
%
\begin{itemize}
\item {\textit{Token embedding}}
\cite{gunel2019sakg,liu2019saeng,jihaozhe2020sakg,yuan2020sesg,huang2020makg,wang2020mesg,liu2021kgbart,ji2021sakg,pasunuru2021masg,zhou2021masg} which is generally produced from input tokens by the last layer of the pretrained language model (e.g., Bidirectional Encoder Representations from Transformers (BERT)). The \textit{WordPiece embedding} \cite{zhang2019sang} is a special token embedding obtained by \textit{WordPiece} tokenizers.
\item {\textit{Word embedding}} \cite{han2016sekg,takase2016sasg,tan2017saug,zhang2017seng,guan2019sakg,koncelkedziorsk2019sakg,fan2019makg,zheng2019meng,zhang2019sang,jihaozhe2020sakg,jin2020sasg,you2021sang} which is typically denoted as a vector of low dimension real numbers via methods, such as the one-hot vector and distributed representation. The \textit{Word2Vec} is a general word embedding algorithm, producing the \textit{Word2Vec embedding} \cite{yasunaga2017mesg,anhdt2019sang}. Also, the word vector garnered by the \textit{Global Vectors ForWord Representation (GloVe)} algorithm and the \textit{FastText} mechanism is known as the \textit{GloVe embedding} \cite{zhang2017seng,tan2017saug,wang2020mesg,ji2021sakg} and the \textit{FastText embedding} \cite{anhdt2019sang}, respectively.
Moreover, the \textit{Context embedding} \cite{takase2016sasg,liu2019saeng} is a contextual vector for output words from the top layer of the language model, mapped with a weight matrix.
%
\item{\textit{Sentence embedding}} \cite{tan2017saug,zheng2019meng,tang2020sekg} which is typically a concatenation of word embeddings or gained by Sent2Vec \cite{zhang2017seng}. In deep neural summarization methods, sentence embeddings are computed from word embeddings \cite{yasunaga2017mesg} or derived by language models (e.g., BERT). Besides, the TF-IDF is a general sentence embedding algorithm for the \textit{TF-IDF embedding} \cite{tang2020sekg}. The term frequency value is neglected in case the summary is formed by tremendously fewer tokens than the source document, known as the \textit{IDF-weighted word embedding} \cite{you2021sang}. Also, an \textit{Elementary Discourse Unit (EDU)} is a sub-sentence phrase unit originating from RST discourse trees, represented by the \textit{EDU embedding} \cite{xu2020sedg}. The \textit{Phrase embedding} \cite{koncelkedziorsk2019sakg} is a special sentence embedding produced from word embeddings run over last hidden states of neural networks (e.g., RNN). The \textit{Title embedding} \cite{koncelkedziorsk2019sakg} is the title word embedding, regarding the title as a sentence, produced by neural networks (e.g., RNN) with last hidden states. 
\item{\textit{Document embedding}} \cite{yasunaga2017mesg,zheng2019meng,zhang2019sang} which is the concatenation of sentence embeddings or computed from sentence embeddings by the neural model.
\item{\textit{Cluster embedding}} \cite{yasunaga2017mesg} which is resulted from averaging document embeddings, supplied in the form of real numbers.

\end{itemize}

\vspace{1.5mm}
\noindent \textbf{Graph Embedding.} Graph embedding methods can be applied on embedding different components of the graph. 

\begin{itemize}
\item {\textit{Node embedding}} \cite{liu2015sasg,takase2016sasg,jin2020sasg,liu2021kgbart,zhu2021sakg} which represents a graph node, computed by the network layer from aggregated local graph information of its adjacent nodes and relations. In terms of the node orientation, the node embedding can be further classified into \textit{Forward-looking node embedding} and \textit{Backward-looking node embedding} \cite{koncelkedziorsk2019sakg}.
\item {\textit{Entity embedding}} \cite{gunel2019sakg,zhou2021masg} which is a representation of a graph entity, learned from output vectors of pretrained language models  or by techniques for modelling multi-relational data, such as the TransE \cite{bordes2013transe}.
\item {\textit{Edge embedding}} \cite{takase2016sasg} which is the representation of an out-edge of the graph directed to the local parent node or global root node.
\item {\textit{Relation embedding}} \cite{liu2015sasg,gunel2019sakg,jihaozhe2020sakg,liu2021kgbart} which is the representation of a relationship or concept between entities, typically derived by TransE from the graph and known as the \textit{Concept embedding} \cite{liu2021kgbart}. Besides, it can be captured by firstly aggregating node and edge embeddings and then transforming it via linear transformations followed by nonlinear activation functions (e.g., ReLU)  \cite{jin2020sasg}. Also, the relationship type can be indicated by the \textit{Relation-type embedding} \cite{jihaozhe2020sakg}.
\item {\textit{Graph weight embedding}} \cite{fan2019makg} which is capable to represent the weight of both the node and edge of a graph, learned from the gating function or discretization of real numbers. A graph weight embedding that sorely indicates the edge weight is represented as a token embedding and is equal to the number of merge operations increased by one, known as the \textit{Edge weight embedding} \cite{wang2020mesg,zhou2021masg}.
\end{itemize}

\vspace{1.5mm}
\noindent \textbf{Topic Embedding.}
A \textit{Topic embedding} \cite{zheng2019meng,cui2020seug} is applied to embed topical information. It is a topic word vector typically composited from document embeddings or distilled subtopics. In deep learning architectures, it can be learned by neural topic models. Besides, the \textit{subtopic embedding} \cite{zheng2019meng} is constructed by sentence embeddings.

\vspace{1.5mm}
\noindent \textbf{Position Embedding.} Position embedding is related to native knowledge. Its embedding is generated straightforward by using the token index information.
%
\begin{itemize}
\item {\textit{Hard-position embedding}} \cite{fan2019makg,liu2019saeng,yuan2020sesg,pasunuru2021masg,wu2021masg} which is the numeric index of a token in its corresponding token sequence (i.e., sentence), also known as the \textit{Positional embedding} \cite{zhang2020mang,jihaozhe2020sakg,jin2020sasg}.
\item {\textit{Soft-position embedding}} \cite{liu2021kgbart} which is the token index in a token sequence tree (i.e., sentence tree), represented as an integer number.
\item {\textit{Segment embedding}} \cite{liu2019saeng,yuan2020sesg} which is a token notation assigned for discriminating multiple adjacent granularity levels (e.g., sentences) in a document, based on the parity of the level index.
%
\end{itemize}

\section{Knowledge Embedding in Different Learning Architectures}
\label{sec:kelearnarc}

\begin{figure*}[ht]
  \centering
  \includegraphics[width=1\textwidth]{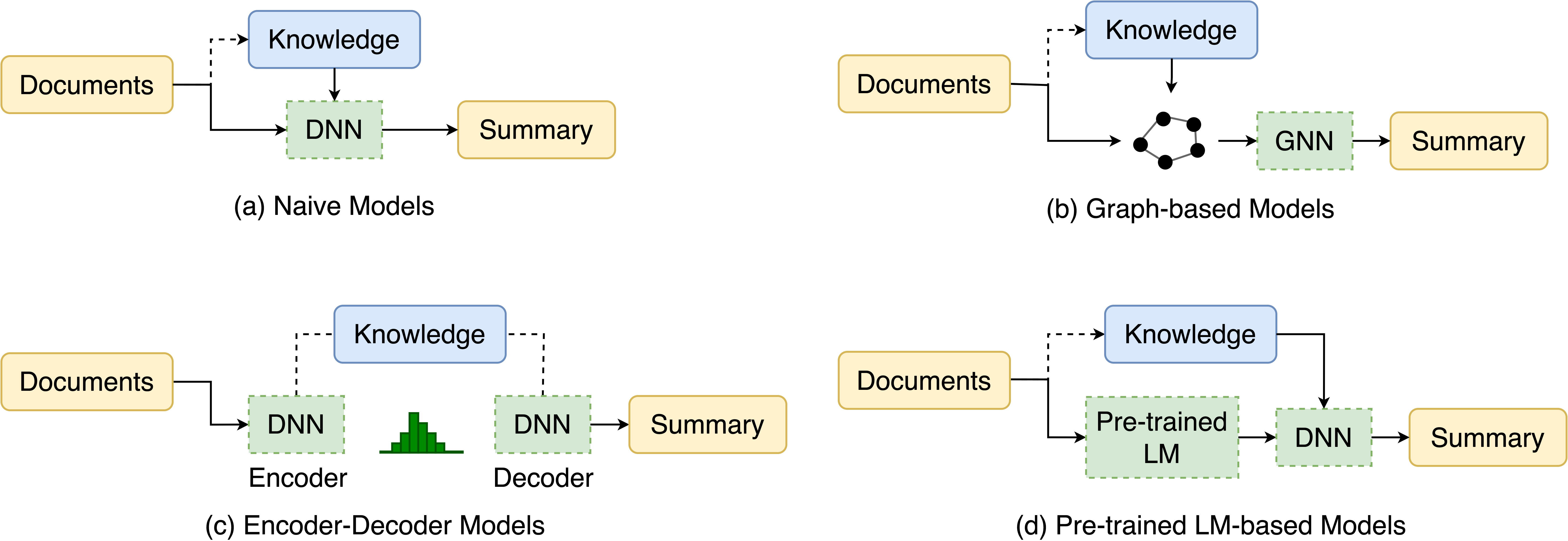}
  \caption{Learning architectures of embedding knowledge for document summarization. The dot line connected to knowledge indicates the knowledge could be obtained from the documents or external sources.}
  \label{fig:dsarchi}
\end{figure*}

In this section, we discuss the reviewed works from the perspective of learning architectures applied for incorporating knowledge in document summarization models, with more attention to deep learning architectures.  Figure  \ref{fig:dsarchi} depicts the four deep learning based architectures to embed knowledge into the summarizers. \textit{Native models} (Figure  \ref{fig:dsarchi} (a)) represent the works that only apply single Deep Neural Networks (DNN) for embedding knowledge. \textit{Graph-based model}  (Figure \ref{fig:dsarchi} (b)) utilize the graph-based neural networks to first form knowledge into graph, then learn from the graphs with graph neural networks (GNN), graph convolutional networks (GCN) or graph attention network (GAT). 
\textit{Encoder-decoder models} (Figure  \ref{fig:dsarchi} (c)) describe the works employ the encoder-decoder architecture and with different implementations for encoder and decoder. \textit{Pre-trained Language model-based models} (Figure  \ref{fig:dsarchi} (d)) introduce the works that are build upon the pre-trained representations.   
Additionally, we combine with the overview in previous sections and provide a summary for the reviewed works, as shown in Table \ref{tab:dsembeddings}, that covers main learning architectures for embeddings, types of document summarization tasks, and types of knowledge embeddings. We particularly highlight specific technique characteristics deriving knowledge embeddings for document summarization.

\subsection{Naive Approaches}
Models for document summarization can employ DNN directly for embedding the knowledge. 
\citet{takase2016sasg} adopted structural syntactic and semantic information as knowledge and utilized a variant of child-sum Tree-LSTM to encode the syntactic and semantic information into fixed-length embedding.
However, due to the advancement of deep learning architectures, in recent years, most works employ more complex architectures rather than simply utilizing only one deep neural network.

%
%

\begin{table*} \scriptsize
\centering
\begin{tabular}{llcl}
\hline

\textbf{\makecell{Learning}} &
\multicolumn{1}{c}{ \multirow{2}{*}{\textbf{Methods}} } &
\textbf{\makecell{DS}} &
\textbf{\makecell{Knowledge}} \\

\multicolumn{1}{c}{\textbf{Architectures}} &
&
\multicolumn{1}{c}{\textbf{Tasks}} &
\multicolumn{1}{c}{\textbf{Embeddings}} \\

\hline

Transformer & SemSUM \cite{jin2020sasg} & A - S  &  TOK, SEN, N, EDG, R, P \\
            & Seq2Seq+MTG \cite{fan2019makg} & A - M  &  W, GW, P  \\
            & PEGASUS \cite{zhang2020mang} & A - M  &  P   \\
BART        & BartGraphSumm \cite{pasunuru2021masg} & A - M  &  TOK, P   \\
BERT        & BERTSUM \cite{liu2019saeng} & A/E - S  &  TOK, W, CON, SEN, D, P, SEG   \\
            & HIBERT \cite{zhang2019seng} & E - S  &  W, SEN, D, P    \\
            & BERT+HGM \cite{yuan2020sesg} & E - S  &  TOK, SEN, D, P, SEG \\
            & RST+spanBERT \cite{huang2021sedg} & E - S  &  SEN, EDU, D, N, ENT, R \\
\textbf{    }\textbf{    }\textbf{    }    + WordPiece  & BERT+RL \cite{zhang2019sang} & A - S  &  WP, W, D   \\
\textbf{    }\textbf{    }\textbf{    }    + GAT & Topic-GraphSum \cite{cui2020seug} & E - S  &  SEN, N, TOP \\
RoBERTa     & ASGARD \cite{huang2020makg} & A - M  &  TOK, N   \\
            & EMSUM \cite{zhou2021masg} & A - M  &  TOK, ENT, EW \\
            & BASS \cite{wu2021masg} & A - S/M  &  TOK, N, P   \\

\hline

RNN         & GraphWriter \cite{koncelkedziorsk2019sakg} & A - S  &  W, SEN, TIT, N, ENT, R   \\
\textbf{    }\textbf{    }\textbf{    }    + W2V   & STDS \cite{zheng2019meng} & E - M  &  W, W2V, SEN,              D, TOP   \\
\textbf{    }\textbf{    }\textbf{    }    + GCN   & GRU+GCN \cite{yasunaga2017mesg} & E - M  &  W, W2V, SEN, D, CLU   \\
LSTM        & PGN+IDF \cite{you2021sang} & A - S  &  W, IDF   \\
\textbf{    }\textbf{    }\textbf{    }    + W2V / FT & PG+PreTrained \cite{anhdt2019sang} & A - S  &  W2V / FT   \\
\textbf{    }\textbf{    }\textbf{    }    + GloVe & RNN+LSTM \cite{zhang2017seng} & E - S  &  W, GV, SEN                \\
TreeLSTM    & ABS+AMR \cite{takase2016sasg} & A - S  &  W, CON, N, EDG \\

\hline

GCN         & GRF \cite{jihaozhe2020sakg} & A - S  &  TOK, W, N, R, P    \\
\textbf{    }\textbf{    }\textbf{    }    + SpanExt  & DISCOBERT \cite{xu2020sedg} & E - S  &  TOK, SEN, EDU    \\
GAT         & HETERSUMGRAPH \cite{wang2020mesg} & E - S/M  &  W, GV, SEN, D, N, EW \\
            & FASUM \cite{zhu2021sakg} & A - S  &  TOK, N \\
\textbf{    }\textbf{    }\textbf{    }    + Glove & SKGSUM \cite{ji2021sakg} & A - S  &  TOK, GV, SEN, TF-IDF, N \\

\hline
Word Rep.   & IE+MSA \cite{guan2019sakg} & A - S  &  CON, ENT, R \\
            & FSGM \cite{han2016sekg} & E - S  &  W, SEN, D   \\
\textbf{    }\textbf{    }\textbf{    }    + Glove & GAM+HBS \cite{tan2017saug} & A - S  &  W, GV, SEN, D \\
TransE      & TXL+WikiKG \cite{gunel2019sakg} & A - S  &  TOK, ENT, R                \\
TF - IDF      & kg-KMTR \cite{tang2020sekg} & E - S  &              SEN, TF-IDF \\

\hline
\end{tabular}
\caption{List of the representative \textbf{A}bstract or \textbf{E}xtractive \textbf{S}ingle- or \textbf{M}ulti-document summarization (\textbf{DS}) methods incorporated the knowledge, indicating knowledge embedding learning architectures. Embedding kinds of
\textbf{TOK}en, WordPiece (\textbf{WP}), 
\textbf{W}ord, Word2Vec (\textbf{W2V}), GloVe (\textbf{GV}), FastText (\textbf{FT}), \textbf{CON}text, 
\textbf{SEN}tence, \textbf{TF-IDF}, \textbf{IDF}, \textbf{EDU}, \textbf{TIT}le, 
\textbf{D}ocument, \textbf{CLU}ster, 
\textbf{N}ode, \textbf{ENT}ity, \textbf{EDG}e, \textbf{R}elation, Graph Weight (\textbf{GW}), Edge Weight (\textbf{EW}), 
\textbf{TOP}ic, 
\textbf{P}osition, and \textbf{SEG}ment are presented.
%
%
}
\label{tab:dsembeddings}
\end{table*}

\subsection{Graph-based Approaches}
The \textit{graph convolutional network} is a novel knowledge embedding approach in document summarization, majoring to embed graph-formed knowledge, such as the knowledge graph ConcepNet5 \cite{jihaozhe2020sakg,xu2020sedg}. As an upgrade from GCN, \textit{graph attention network} is widely utilized in document summarization for embedding knowledge extracted by OpenIE or Stanford CoreNLP, preserved in graphs \cite{zhu2021sakg,ji2021sakg}. 
%
%
In addition, \citet{wang2020mesg} considered heterogeneous word-sentence relations to preserve hierarchical information as knowledge. The model treats the whole document as a graph and uses a graph attention network to learn the embedding. The knowledge-embedded sentences representations of sentence nodes is finally used for summary selection.
Moreover, GraphWriter \cite{koncelkedziorsk2019sakg} employs the \textit{Science-domain Information Extraction (SciIE)} for extracting science knowledge and embedding the knowledge by graph attention network.

\subsection{Encoder-Decoder based Approaches}
In this category, the encoder-decoder architecture is adopted for  document summarization. 
The knowledge could be either embedded in encoder or decoder or both. 
%
%
EMSUM \cite{zhou2021masg} uses the RoBERTa to embed the documentary information extracted by the \textit{Coreference Resolution Tool} from \textit{AllenNLP}. The Transformer-based encoder-decoder framework with a heterogeneous graph consisting of text units and entities as nodes.
\citet{zhu2021sakg} applied a Transformer-based encoder-decoder architecture via attention. The knowledge graph is obtained from information extraction results and participates in the decoder’s attention.
\citet{zheng2019meng} adopted a bidirectional RNN encoder-decoder framework to learn sentence embedding for leveraging topic knowledge. Topic embedding is learnt from a soft-clustering on the sentence embedding and is fused with the model by non-linear transformation to the encoder.
SemSUM \cite{jin2020sasg} employs the Transformer encoder in a Transformer-based encoder-decoder model to encode knowledge and learn knowledge embeddings from the syntactic knowledge extracted by an off-the-shelf dependency parser.
\citet{wu2021masg} utilized the semantic graph via dependency parsing and encoded knowledge in both encoder and decoder in a Transformer-based encoder-decoder model.
\citet{ji2021sakg} experimented with three types of knowledge constructing three knowledge graphs: entity graph, similarity graph, and discourse graph, and encoded the knowledge graph in both encoder and decoder of a Transformer-based encoder-decoder architecture.
The entity graph is obtained by applying \textit{Named Entity Recognition (NER)} and open information extraction via third part tools. The similarity graph presents the sentence cosine similarity by using TF-IDF vectorization. The discourse graph follows the way to build an \textit{Approximate Discourse Graph (ADG)} \cite{yasunaga2017mesg}.

\subsection{Pre-trained Model-based Approaches} 

\citet{yuan2020sesg} obtained informative sentence representations on BERT, containing a hierarchical graph that brings the tokens at each granularity level be able to capture semantics from different sources.
\citet{xu2020sedg} leveraged discourse knowledge within structural discourse graphs constructed based on RST trees and coreference mentions. In addition, the work encodes the discourse graph using graph convolutional networks, which serve as graph encoders based upon sentence representations from BERT. The encoder finally outputs the knowledge incorporated embedding into MLP layers.
\citet{cui2020seug} utilized topic knowledge via neural topic model and built a heterogeneous document graph consisting of sentence and topic nodes to learn the representations by a modified graph attention network with BERT. The representations of sentence nodes are extracted to compute the final summary.
BartGraphSumm \cite{pasunuru2021masg} 
is equipped with BART to encode the semantic knowledge extracted by the OpenIE from documents.  Moreover, the model relies on the pre-trained RoBERTa encoder, which uses GAT to learn the graph extracted by OpenIE.

\subsection{Non-deep learning Approaches.}
In contrast to DNN-based approaches, non-deep learning approaches are commonly adopted for embedding external knowledge into document summarization. 
The traditional word vector representation method is still utilized in recent knowledge embedding approaches for document summarization \cite{guan2019sakg}. Besides, advanced word vector representation methods, such as the distributed representation, are applied in both abstractive and extractive document summarization tasks, representing each word by its distributed representation \cite{han2016sekg,tan2017saug}.
In addition, some document summarization knowledge embedding approaches adopt the linearization mechanism (e.g., TransE) to linearize the knowledge into sequences for embedding into the model architecture \cite{gunel2019sakg}. 
Scarce approaches utilize straight the embedding algorithms, such as the TF-IDF algorithm, for learning knowledge embeddings in document summarization models \cite{tang2020sekg}.

\section{Challenges and Future Opportunities}
\label{sec:outlook}
As still in its evolutionary stage, the research of embedding knowledge into document summarization faces numerous challenges and remains unfilled gaps. 
In this section, we discuss the challenges and promising avenues of ongoing and future works aligned with the goals \textbf{G1} to \textbf{G5}.

\subsection{Knowledge Quality}
For document summarization models that leverage knowledge, the knowledge that covers less information, retains fault information, or contains factual errors can significantly harm the summarization performance. There are some latent future directions to maintaining the quality of a knowledge base for document summarization to carry large amounts of essential and factually consistent information.

\paragraph{Knowledge Collection} The issue of fact coverage can occur due to the choice of information extraction strategies when collecting knowledge from texts. This issue may cause the lost of prominent information from source documents
, thus degrading the quality of the generated summary \cite{koncelkedziorsk2019sakg}.
Therefore, an effective extraction strategy designed for improving the coverage in knowledge collection is requisite to be explored, in order to reduce the missing knowledge in the distilling process. It can also further help produce informative summaries corresponding to the goals \textbf{G1} and \textbf{G2}.
Besides, it is noted that determining the voice (i.e., active or passive) of sentences while extracting factual triples from the source text can advance the extracted information quality \cite{abdi2017mekg}, which helps avoid missing information.
More future research can be put in this direction to ensure the quality of collected knowledge.

\paragraph{Knowledge Purification} The knowledge graph entity disambiguation is to conquer entity ambiguity problems by matching ambiguous text entities to the corresponding knowledge graph entities for knowledge purification.
However, when operating the disambiguation process to the knowledge graph to eliminate redundancy knowledge for document summarization, some of the salient text information could be lost \cite{gunel2019sakg}. Thus, a better strategy for knowledge graph disambiguation to condense the summary while remaining primary contents can be necessary.
Also, more effective mechanisms for the entity recognition and entity linking within a knowledge graph to maintain more relations of knowledge entities are worth investigating \cite{tang2020sekg}. Entity linking herein is the process of linking textual mentions of entities from source texts to the corresponding entities in a knowledge graph.
These mechanisms can better reduce redundant information while retaining the knowledge base quality, achieving the goal \textbf{G2}.

\paragraph{Knowledge Consistency} Factual inconsistency errors refer to fact conflicts, categorized into contradicted fact (i.e., intrinsic error) and irrelevant fact (i.e., extrinsic error) to source text facts \cite{xie2021inconerror}.
The knowledge from open-source knowledge graphs or extracted from the source corpora can inevitably involve varied intrinsic or extrinsic errors. Seriously erroneous knowledge can harm a knowledge-embedded summarizer's performance terribly, mostly clashing with the goal \textbf{G5}.
Even if factual inconsistency errors have been recognized and attached importance, scarce studies are qualified to precisely address and tackle inconsistency errors in document summarization. It is because inconsistency errors can be detected hardly by linguistic analysis. And since the knowledge databases for document summarization tasks are generally large-scale, it can cost laborious efforts to check each knowledge entity relation. Thus, exploring advanced ways to scan and solve inconsistent errors for incorporated external or personalized knowledge graphs, which contain plenty of mixed facts, can be a valuable future direction for knowledge-based document summarization.

\subsection{Knowledge from Multi-facets}
Except for investigations of manners to retain most of the superior knowledge, fusing knowledge from multiple facets to enhance the knowledge coverage in terms of different aspects can also promote the goals \textbf{G1} to \textbf{G5} for document summarization tasks.

\paragraph{Knowledge from Multiple Resources} 
Incorporating multiple types of knowledge from the real world and source corpora can be more beneficial in gathering more facts and prominent information in document summarization. It can improve the fact consistency of generated informative summaries, achieving the goals \textbf{G1} and \textbf{G5}. As aforementioned in Section \ref{sec:knowtaxo}, recent document summarization studies reported the possibility and advantages of blending different kinds of knowledge, expanding the knowledge base and enhancing the commonsense uniformity. Also, integrating knowledge graphs with abstract meaning representations to combine knowledge for document summarization can be an appealing research direction \cite{jin2020sasg}. However, empirically verifying the efficiency of leveraging fused knowledge and exploring effective multiple knowledge combinations are not well-attended research areas.

\paragraph{Knowledge from Multiple Levels} 
Recent works reported that sentence-level or paragraph-level relation extraction methods might lose global relationship information from the entire document context \cite{jain2020scirex}. The lack of high-level relations, e.g., relations among paragraphs, can mainly reduce the summary coherence and harm the goal \textbf{G3}. However, most of current knowledge extraction methods for document summarization still focus on firstly splitting the entire document into sentences and then extracting triples from the sentence span as the sentence-level relation extraction, which loses varying degrees of the context information and high-level knowledge. Thus, it inspires the research direction of 
extracting document-level knowledge for document summarization. 
Moreover, employing the novel text-to-graph summarizer with the knowledge graph usage to capture more multi-level relations among knowledge can be a promising research direction in document summarization \cite{wu2020kgie}.

\subsection{Knowledge Embedding Techniques}
As summarized in Section \ref{sec:kelearnarc}, the majority of recent document summarization works utilized encoder-decoder models, neural networks, and non-deep learning knowledge embedding methods to generate embeddings when incorporating the knowledge into models. Explorations and experiments of employing novel knowledge embedding approaches, which widely benefit natural language processing tasks not limited to document summarization tasks, to achieve the goals \textbf{G1} to \textbf{G5} can be worthwhile future works. 

\paragraph{Novel Knowledge Embedding Methods} The \textit{FocusE} \cite{pai2021focuse} enhances the knowledge preservation by forming and merging textual information in numerical forms with lexical knowledge  
while remaining the textual 
information. 
%
It presents the potential of combining multiple forms of knowledge within one knowledge graph for further better knowledge embedding, improving the fact richness and accuracy for the goals \textbf{G1} to \textbf{G5}. 
Future research could adopt the idea and explore the ways to jointly embed knowledge from heterogeneous sources and to different forms.  

\paragraph{Novel Learning Strategies} As discussed in Section \ref{sec:kelearnarc}, the mostly used learning architectures in the reviewed works are graph-based, encode-decoder and pre-trained models. There is a space to explore more learning architectures that could help achieve the goals \textbf{G1} to \textbf{G5} by incorporating knowledge.  
Reinforcement learning has been applied in document summarization as a part of the overall model to train the model by giving rewards. However, most of works utilize the evaluation metrics as the rewards. Few works consider the knowledge as part of the rewards  \cite{sharma2019sang}. Future works could investigate how to form the informative knowledge as rewards to train the summarizers. 
Besides, the recently popular \textit{Prompting} \cite{liu2021pre} strategy promotes another way of adopting pretrained language models. It has not yet been well-adopted in summarization community.  Exploring the methods of embedding knowledge for document summarization via prompting strategy is another promising future direction.


\section{Conclusion}
\label{sec:conclude}
Along with the pursuit of more informative and coherent summaries with factual consistency, attention to knowledge embedding as an incorporation module for document summarizers to enhance model performance and improve summary quality gathered pace. In this paper, we surveyed the state-of-the-art approaches to embedding knowledge into document summarization models. To explicitly review each representative knowledge embedding approach in document summarization, we proposed taxonomies for knowledge and knowledge embeddings and explored embedding learning architectures under the document summarization perspective. Furthermore, we discussed open questions and appealing research directions for embedding knowledge in document summarization tasks, which we hope can drive new improvements in the document summarization field.

\bibliography{mybibfile}

\begin{thebibliography}{64}
\expandafter\ifx\csname natexlab\endcsname\relax\def\natexlab#1{#1}\fi
\providecommand{\url}[1]{\texttt{#1}}
\providecommand{\href}[2]{#2}
\providecommand{\path}[1]{#1}
\providecommand{\DOIprefix}{doi:}
\providecommand{\ArXivprefix}{arXiv:}
\providecommand{\URLprefix}{URL: }
\providecommand{\Pubmedprefix}{pmid:}
\providecommand{\doi}[1]{\href{http://dx.doi.org/#1}{\path{#1}}}
\providecommand{\Pubmed}[1]{\href{pmid:#1}{\path{#1}}}
\providecommand{\bibinfo}[2]{#2}
\ifx\xfnm\relax \def\xfnm[#1]{\unskip,\space#1}\fi
\bibitem[{Ma et~al.(2022)Ma, Zhang, Guo, Wang, and Sheng}]{ma2020MDSsur}
\bibinfo{author}{C.~Ma}, \bibinfo{author}{W.~E. Zhang},
  \bibinfo{author}{M.~Guo}, \bibinfo{author}{H.~Wang}, \bibinfo{author}{Q.~Z.
  Sheng},
\newblock \bibinfo{title}{{Multi-document Summarization via Deep Learning
  Techniques: A Survey}},
\newblock \bibinfo{journal}{ACM Computing Surveys}  (\bibinfo{year}{2022})
  \bibinfo{pages}{1--35}. \DOIprefix\doi{10.1145/3529754}.
\bibitem[{El-Kassas et~al.(2021)El-Kassas, Salama, Rafea, and
  Mohamed}]{elkassas2021DSsur}
\bibinfo{author}{W.~S. El-Kassas}, \bibinfo{author}{C.~R. Salama},
  \bibinfo{author}{A.~A. Rafea}, \bibinfo{author}{H.~K. Mohamed},
\newblock \bibinfo{title}{Automatic text summarization: A comprehensive
  survey},
\newblock \bibinfo{journal}{ESWA} \bibinfo{volume}{165} (\bibinfo{year}{2021})
  \bibinfo{pages}{113679}. \DOIprefix\doi{10.1016/j.eswa.2020.113679}.
\bibitem[{Ferreira et~al.(2014)Ferreira, de~Souza~Cabral, de~Freitas, Lins,
  e~Silva, Simske, and Favaro}]{ferreira2014MDSsur}
\bibinfo{author}{R.~Ferreira}, \bibinfo{author}{L.~de~Souza~Cabral},
  \bibinfo{author}{F.~L.~G. de~Freitas}, \bibinfo{author}{R.~D. Lins},
  \bibinfo{author}{G.~P. e~Silva}, \bibinfo{author}{S.~J. Simske},
  \bibinfo{author}{L.~Favaro},
\newblock \bibinfo{title}{A multi-document summarization system based on
  statistics and linguistic treatment},
\newblock \bibinfo{journal}{ESWA} \bibinfo{volume}{41} (\bibinfo{year}{2014})
  \bibinfo{pages}{5780--5787}. \DOIprefix\doi{10.1016/j.eswa.2014.03.023}.
\bibitem[{Hogan et~al.(2022)Hogan, Blomqvist, Cochez, d’Amato, Melo,
  Gutierrez, Kirrane, Gayo, Navigli, Neumaier, Ngomo, Polleres, Rashid, Rula,
  Schmelzeisen, Sequeda, Staab, and Zimmermann}]{hogan2021kgesur}
\bibinfo{author}{A.~Hogan}, \bibinfo{author}{E.~Blomqvist},
  \bibinfo{author}{M.~Cochez}, \bibinfo{author}{C.~d’Amato},
  \bibinfo{author}{G.~D. Melo}, \bibinfo{author}{C.~Gutierrez},
  \bibinfo{author}{S.~Kirrane}, \bibinfo{author}{J.~E.~L. Gayo},
  \bibinfo{author}{R.~Navigli}, \bibinfo{author}{S.~Neumaier},
  \bibinfo{author}{A.-C.~N. Ngomo}, \bibinfo{author}{A.~Polleres},
  \bibinfo{author}{S.~M. Rashid}, \bibinfo{author}{A.~Rula},
  \bibinfo{author}{L.~Schmelzeisen}, \bibinfo{author}{J.~Sequeda},
  \bibinfo{author}{S.~Staab}, \bibinfo{author}{A.~Zimmermann},
\newblock \bibinfo{title}{Knowledge graphs},
\newblock \bibinfo{journal}{ACM Computing Surveys} \bibinfo{volume}{54}
  (\bibinfo{year}{2022}) \bibinfo{pages}{1--37}.
  \DOIprefix\doi{10.1145/3447772}.
\bibitem[{Gao et~al.(2021)Gao, Chen, Ren, Zhao, and Yan}]{newkninDS2021}
\bibinfo{author}{S.~Gao}, \bibinfo{author}{X.~Chen}, \bibinfo{author}{Z.~Ren},
  \bibinfo{author}{D.~Zhao}, \bibinfo{author}{R.~Yan},
\newblock \bibinfo{title}{{From Standard Summarization to New Tasks and Beyond:
  Summarization with Manifold Information}},
\newblock in: \bibinfo{booktitle}{IJCAI}, \bibinfo{year}{2021}, pp.
  \bibinfo{pages}{4854--4860}. \DOIprefix\doi{10.48550/arXiv.2005.04684}.
\bibitem[{Han et~al.(2016)Han, Lv, Hu, Wang, and Wang}]{han2016sekg}
\bibinfo{author}{X.~Han}, \bibinfo{author}{T.~Lv}, \bibinfo{author}{Z.~Hu},
  \bibinfo{author}{X.~Wang}, \bibinfo{author}{C.~Wang},
\newblock \bibinfo{title}{{Text Summarization Using {FrameNet}-Based Semantic
  Graph Model}},
\newblock \bibinfo{journal}{Scientific Programming} \bibinfo{volume}{2016}
  (\bibinfo{year}{2016}) \bibinfo{pages}{1--10}.
  \DOIprefix\doi{10.1155/2016/5130603}.
\bibitem[{Yasunaga et~al.(2017)Yasunaga, Zhang, Meelu, Pareek, Srinivasan, and
  Radev}]{yasunaga2017mesg}
\bibinfo{author}{M.~Yasunaga}, \bibinfo{author}{R.~Zhang},
  \bibinfo{author}{K.~Meelu}, \bibinfo{author}{A.~Pareek},
  \bibinfo{author}{K.~Srinivasan}, \bibinfo{author}{D.~Radev},
\newblock \bibinfo{title}{{Graph-based Neural Multi-Document Summarization}},
\newblock in: \bibinfo{booktitle}{CoNLL}, \bibinfo{year}{2017}, pp.
  \bibinfo{pages}{452--462}. \DOIprefix\doi{10.18653/v1/K17-1045}.
\bibitem[{Tan et~al.(2017)Tan, Wan, and Xiao}]{tan2017saug}
\bibinfo{author}{J.~Tan}, \bibinfo{author}{X.~Wan}, \bibinfo{author}{J.~Xiao},
\newblock \bibinfo{title}{{Abstractive Document Summarization with a
  Graph-Based Attentional Neural Model}},
\newblock in: \bibinfo{booktitle}{ACL}, \bibinfo{year}{2017}, pp.
  \bibinfo{pages}{1171--1181}. \DOIprefix\doi{10.18653/v1/P17-1108}.
\bibitem[{Gunel et~al.(2019)Gunel, Zhu, Zeng, and Huang}]{gunel2019sakg}
\bibinfo{author}{B.~Gunel}, \bibinfo{author}{C.~Zhu},
  \bibinfo{author}{M.~Zeng}, \bibinfo{author}{X.~Huang},
\newblock \bibinfo{title}{{Mind The Facts: Knowledge-Boosted Coherent
  Abstractive Text Summarization}},
\newblock in: \bibinfo{booktitle}{NeurIPS}, \bibinfo{year}{2019}, pp.
  \bibinfo{pages}{1--7}. \DOIprefix\doi{10.48550/arXiv.2006.15435}.
\bibitem[{Ji and Zhao(2021)}]{ji2021sakg}
\bibinfo{author}{X.~Ji}, \bibinfo{author}{W.~Zhao},
\newblock \bibinfo{title}{{SKGSUM: Abstractive Document Summarization with
  Semantic Knowledge Graphs}},
\newblock in: \bibinfo{booktitle}{IJCNN}, \bibinfo{year}{2021}, pp.
  \bibinfo{pages}{1--8}. \DOIprefix\doi{10.1109/IJCNN52387.2021.9533494}.
\bibitem[{Tang et~al.(2020)Tang, Yuan, Tang, and Chen}]{tang2020sekg}
\bibinfo{author}{T.~Tang}, \bibinfo{author}{T.~Yuan},
  \bibinfo{author}{X.~Tang}, \bibinfo{author}{D.~Chen},
\newblock \bibinfo{title}{{Incorporating External Knowledge into Unsupervised
  Graph Model for Document Summarization}},
\newblock \bibinfo{journal}{Electronics} \bibinfo{volume}{9}
  (\bibinfo{year}{2020}) \bibinfo{pages}{1--13}.
  \DOIprefix\doi{10.3390/electronics9091520}.
\bibitem[{Wu et~al.(2021)Wu, Li, Xiao, Liu, Cao, Li, Wu, and Wang}]{wu2021masg}
\bibinfo{author}{W.~Wu}, \bibinfo{author}{W.~Li}, \bibinfo{author}{X.~Xiao},
  \bibinfo{author}{J.~Liu}, \bibinfo{author}{Z.~Cao}, \bibinfo{author}{S.~Li},
  \bibinfo{author}{H.~Wu}, \bibinfo{author}{H.~Wang},
\newblock \bibinfo{title}{{BASS: Boosting Abstractive Summarization with
  Unified Semantic Graph}},
\newblock in: \bibinfo{booktitle}{ACL-IJCNLP}, \bibinfo{year}{2021}, pp.
  \bibinfo{pages}{6052--6067}. \DOIprefix\doi{10.18653/v1/2021.acl-long.472}.
\bibitem[{Chen et~al.(2021)Chen, Li, Liu, Xiao, Wu, and Wang}]{chen2021mesdg}
\bibinfo{author}{M.~Chen}, \bibinfo{author}{W.~Li}, \bibinfo{author}{J.~Liu},
  \bibinfo{author}{X.~Xiao}, \bibinfo{author}{H.~Wu},
  \bibinfo{author}{H.~Wang},
\newblock \bibinfo{title}{{SgSum: Transforming Multi-document Summarization
  into Sub-graph Selection}},
\newblock in: \bibinfo{booktitle}{EMNLP}, \bibinfo{year}{2021}, pp.
  \bibinfo{pages}{4063--4074}. \DOIprefix\doi{10.18653/v1/2021.emnlp-main.333}.
\bibitem[{Wang et~al.(2017)Wang, Mao, Wang, and Guo}]{wang2017kgesur}
\bibinfo{author}{Q.~Wang}, \bibinfo{author}{Z.~Mao}, \bibinfo{author}{B.~Wang},
  \bibinfo{author}{L.~Guo},
\newblock \bibinfo{title}{{Knowledge Graph Embedding: A Survey of Approaches
  and Applications}},
\newblock \bibinfo{journal}{IEEE TKDE} \bibinfo{volume}{29}
  (\bibinfo{year}{2017}) \bibinfo{pages}{2724--2743}.
  \DOIprefix\doi{10.1109/TKDE.2017.2754499}.
\bibitem[{Cai et~al.(2018)Cai, Zheng, and Chang}]{cai2018gesur}
\bibinfo{author}{H.~Cai}, \bibinfo{author}{V.~W. Zheng},
  \bibinfo{author}{K.~C.-C. Chang},
\newblock \bibinfo{title}{{A Comprehensive Survey of Graph Embedding: Problems,
  Techniques, and Applications}},
\newblock \bibinfo{journal}{IEEE TKDE} \bibinfo{volume}{30}
  (\bibinfo{year}{2018}) \bibinfo{pages}{1616--1637}.
  \DOIprefix\doi{10.1109/TKDE.2018.2807452}.
\bibitem[{Xu(2021)}]{xu2021gesur}
\bibinfo{author}{M.~Xu},
\newblock \bibinfo{title}{{Understanding Graph Embedding Methods and Their
  Applications}},
\newblock \bibinfo{journal}{SIREV} \bibinfo{volume}{63} (\bibinfo{year}{2021})
  \bibinfo{pages}{825--853}. \DOIprefix\doi{10.1137/20M1386062}.
\bibitem[{Ji et~al.(2022)Ji, Pan, Cambria, Marttinen, and Yu}]{ji2021kgesur}
\bibinfo{author}{S.~Ji}, \bibinfo{author}{S.~Pan},
  \bibinfo{author}{E.~Cambria}, \bibinfo{author}{P.~Marttinen},
  \bibinfo{author}{P.~S. Yu},
\newblock \bibinfo{title}{{A Survey on Knowledge Graphs: Representation,
  Acquisition and Applications}},
\newblock \bibinfo{journal}{IEEE TNNLS} \bibinfo{volume}{33}
  (\bibinfo{year}{2022}) \bibinfo{pages}{494--514}.
  \DOIprefix\doi{10.1109/TNNLS.2021.3070843}.
\bibitem[{Takase et~al.(2016)Takase, Suzuki, Okazaki, Hirao, and
  Nagata}]{takase2016sasg}
\bibinfo{author}{S.~Takase}, \bibinfo{author}{J.~Suzuki},
  \bibinfo{author}{N.~Okazaki}, \bibinfo{author}{T.~Hirao},
  \bibinfo{author}{M.~Nagata},
\newblock \bibinfo{title}{{Neural Headline Generation on Abstract Meaning
  Representation}},
\newblock in: \bibinfo{booktitle}{EMNLP}, \bibinfo{year}{2016}, pp.
  \bibinfo{pages}{1054--1059}. \DOIprefix\doi{10.18653/v1/D16-1112}.
\bibitem[{Liu and Lapata(2019)}]{liu2019saeng}
\bibinfo{author}{Y.~Liu}, \bibinfo{author}{M.~Lapata},
\newblock \bibinfo{title}{{Text Summarization with Pretrained Encoders}},
\newblock in: \bibinfo{booktitle}{EMNLP-IJCNLP}, \bibinfo{year}{2019}, pp.
  \bibinfo{pages}{3730--3740}. \DOIprefix\doi{10.18653/v1/D19-1387}.
\bibitem[{Zhang et~al.(2019)Zhang, Cai, Xu, and Wang}]{zhang2019sang}
\bibinfo{author}{H.~Zhang}, \bibinfo{author}{J.~Cai}, \bibinfo{author}{J.~Xu},
  \bibinfo{author}{J.~Wang},
\newblock \bibinfo{title}{{Pretraining-Based Natural Language Generation for
  Text Summarization}},
\newblock in: \bibinfo{booktitle}{CoNLL}, \bibinfo{year}{2019}, pp.
  \bibinfo{pages}{789--797}. \DOIprefix\doi{10.18653/v1/K19-1074}.
\bibitem[{Koncel-Kedziorski et~al.(2019)Koncel-Kedziorski, Bekal, Luan, Lapata,
  and Hajishirzi}]{koncelkedziorsk2019sakg}
\bibinfo{author}{R.~Koncel-Kedziorski}, \bibinfo{author}{D.~Bekal},
  \bibinfo{author}{Y.~Luan}, \bibinfo{author}{M.~Lapata},
  \bibinfo{author}{H.~Hajishirzi},
\newblock \bibinfo{title}{{Text Generation from Knowledge Graphs with Graph
  Transformers}},
\newblock in: \bibinfo{booktitle}{NAACL-HLT}, \bibinfo{year}{2019}, pp.
  \bibinfo{pages}{2284--2293}. \DOIprefix\doi{10.18653/v1/N19-1238}.
\bibitem[{Anh and Trang(2019)}]{anhdt2019sang}
\bibinfo{author}{D.~T. Anh}, \bibinfo{author}{N.~T.~T. Trang},
\newblock \bibinfo{title}{{Abstractive Text Summarization Using
  Pointer-Generator Networks With Pre-Trained Word Embedding}},
\newblock in: \bibinfo{booktitle}{SoICT}, \bibinfo{year}{2019}, pp.
  \bibinfo{pages}{473--478}. \DOIprefix\doi{10.1145/3368926.3369728}.
\bibitem[{Guan et~al.(2019)Guan, Wang, and Huang}]{guan2019sakg}
\bibinfo{author}{J.~Guan}, \bibinfo{author}{Y.~Wang},
  \bibinfo{author}{M.~Huang},
\newblock \bibinfo{title}{{Story Ending Generation with Incremental Encoding
  and Commonsense Knowledge}},
\newblock in: \bibinfo{booktitle}{AAAI-IAAI-EAAI}, \bibinfo{year}{2019}, pp.
  \bibinfo{pages}{6473--–6480}. \DOIprefix\doi{10.1609/aaai.v33i01.33016473}.
\bibitem[{Jin et~al.(2020)Jin, Wang, and Wan}]{jin2020sasg}
\bibinfo{author}{H.~Jin}, \bibinfo{author}{T.~Wang}, \bibinfo{author}{X.~Wan},
\newblock \bibinfo{title}{{SemSUM: Semantic Dependency Guided Neural
  Abstractive Summarization}},
\newblock in: \bibinfo{booktitle}{AAAI}, \bibinfo{year}{2020}, pp.
  \bibinfo{pages}{8026--8033}. \DOIprefix\doi{10.1609/aaai.v34i05.6312}.
\bibitem[{Ji et~al.(2020)Ji, Ke, Huang, Wei, Zhu, and Huang}]{jihaozhe2020sakg}
\bibinfo{author}{H.~Ji}, \bibinfo{author}{P.~Ke}, \bibinfo{author}{S.~Huang},
  \bibinfo{author}{F.~Wei}, \bibinfo{author}{X.~Zhu},
  \bibinfo{author}{M.~Huang},
\newblock \bibinfo{title}{{Language Generation with Multi-Hop Reasoning on
  Commonsense Knowledge Graph}},
\newblock in: \bibinfo{booktitle}{EMNLP}, \bibinfo{year}{2020}, pp.
  \bibinfo{pages}{725--736}. \DOIprefix\doi{10.18653/v1/2020.emnlp-main.54}.
\bibitem[{You et~al.(2021)You, Hu, Kamigaito, Takamura, and
  Okumura}]{you2021sang}
\bibinfo{author}{J.~You}, \bibinfo{author}{C.~Hu},
  \bibinfo{author}{H.~Kamigaito}, \bibinfo{author}{H.~Takamura},
  \bibinfo{author}{M.~Okumura},
\newblock \bibinfo{title}{{Abstractive Document Summarization with Word
  Embedding Reconstruction}},
\newblock in: \bibinfo{booktitle}{RANLP}, \bibinfo{year}{2021}, pp.
  \bibinfo{pages}{1586--1596}. \DOIprefix\doi{10.26615/978-954-452-072-4_178}.
\bibitem[{Zhu et~al.(2021)Zhu, Hinthorn, Xu, Zeng, Zeng, Huang, and
  Jiang}]{zhu2021sakg}
\bibinfo{author}{C.~Zhu}, \bibinfo{author}{W.~Hinthorn},
  \bibinfo{author}{R.~Xu}, \bibinfo{author}{Q.~Zeng},
  \bibinfo{author}{M.~Zeng}, \bibinfo{author}{X.~Huang},
  \bibinfo{author}{M.~Jiang},
\newblock \bibinfo{title}{{Enhancing Factual Consistency of Abstractive
  Summarization}},
\newblock in: \bibinfo{booktitle}{NAACL-HLT}, \bibinfo{year}{2021}, pp.
  \bibinfo{pages}{718--733}. \DOIprefix\doi{10.18653/v1/2021.naacl-main.58}.
\bibitem[{Fan et~al.(2019)Fan, Gardent, Braud, and Bordes}]{fan2019makg}
\bibinfo{author}{A.~Fan}, \bibinfo{author}{C.~Gardent},
  \bibinfo{author}{C.~Braud}, \bibinfo{author}{A.~Bordes},
\newblock \bibinfo{title}{{Using Local Knowledge Graph Construction to Scale
  Seq2Seq Models to Multi-Document Inputs}},
\newblock in: \bibinfo{booktitle}{EMNLP-IJCNLP}, \bibinfo{year}{2019}, pp.
  \bibinfo{pages}{4186--4196}. \DOIprefix\doi{10.18653/v1/D19-1428}.
\bibitem[{Zhang et~al.(2020)Zhang, Zhao, Saleh, and Liu}]{zhang2020mang}
\bibinfo{author}{J.~Zhang}, \bibinfo{author}{Y.~Zhao},
  \bibinfo{author}{M.~Saleh}, \bibinfo{author}{P.~J. Liu},
\newblock \bibinfo{title}{{PEGASUS: Pre-training with Extracted Gap-sentences
  for Abstractive Summarization}},
\newblock in: \bibinfo{booktitle}{ICML}, \bibinfo{year}{2020}, pp.
  \bibinfo{pages}{11328--11339}. \DOIprefix\doi{10.48550/arXiv.1912.08777}.
\bibitem[{Huang et~al.(2020)Huang, Wu, and Wang}]{huang2020makg}
\bibinfo{author}{L.~Huang}, \bibinfo{author}{L.~Wu}, \bibinfo{author}{L.~Wang},
\newblock \bibinfo{title}{{Knowledge Graph-Augmented Abstractive Summarization
  with Semantic-Driven Cloze Reward}},
\newblock in: \bibinfo{booktitle}{ACL}, \bibinfo{year}{2020}, pp.
  \bibinfo{pages}{5094--5107}. \DOIprefix\doi{10.18653/v1/2020.acl-main.457}.
\bibitem[{Pasunuru et~al.(2021)Pasunuru, Liu, Bansal, Ravi, and
  Dreyer}]{pasunuru2021masg}
\bibinfo{author}{R.~Pasunuru}, \bibinfo{author}{M.~Liu},
  \bibinfo{author}{M.~Bansal}, \bibinfo{author}{S.~Ravi},
  \bibinfo{author}{M.~Dreyer},
\newblock \bibinfo{title}{{Efficiently Summarizing Text and Graph Encodings of
  Multi-Document Clusters}},
\newblock in: \bibinfo{booktitle}{NAACL-HLT}, \bibinfo{year}{2021}, pp.
  \bibinfo{pages}{4768--4779}. \DOIprefix\doi{10.18653/v1/2021.naacl-main.380}.
\bibitem[{Zhou et~al.(2021)Zhou, Ren, Liu, Su, and Lu}]{zhou2021masg}
\bibinfo{author}{H.~Zhou}, \bibinfo{author}{W.~Ren}, \bibinfo{author}{G.~Liu},
  \bibinfo{author}{B.~Su}, \bibinfo{author}{W.~Lu},
\newblock \bibinfo{title}{{Entity-Aware Abstractive Multi-Document
  Summarization}},
\newblock in: \bibinfo{booktitle}{ACL-IJCNLP}, \bibinfo{year}{2021}, pp.
  \bibinfo{pages}{351--362}. \DOIprefix\doi{10.18653/v1/2021.findings-acl.30}.
\bibitem[{Zhang et~al.(2017)Zhang, Sah, Nguyen, Peri, Loui, Salvaggio, and
  Ptucha}]{zhang2017seng}
\bibinfo{author}{C.~Zhang}, \bibinfo{author}{S.~Sah},
  \bibinfo{author}{T.~Nguyen}, \bibinfo{author}{D.~K. Peri},
  \bibinfo{author}{A.~C. Loui}, \bibinfo{author}{C.~Salvaggio},
  \bibinfo{author}{R.~W. Ptucha},
\newblock \bibinfo{title}{{Semantic Sentence Embeddings for Paraphrasing and
  Text Summarization}},
\newblock \bibinfo{journal}{IEEE GlobalSIP}  (\bibinfo{year}{2017})
  \bibinfo{pages}{705--709}. \DOIprefix\doi{10.1109/GlobalSIP.2017.8309051}.
\bibitem[{Zhang et~al.(2019)Zhang, Wei, and Zhou}]{zhang2019seng}
\bibinfo{author}{X.~Zhang}, \bibinfo{author}{F.~Wei},
  \bibinfo{author}{M.~Zhou},
\newblock \bibinfo{title}{{HIBERT: Document Level Pre-training of Hierarchical
  Bidirectional Transformers for Document Summarization}},
\newblock in: \bibinfo{booktitle}{ACL}, \bibinfo{year}{2019}, pp.
  \bibinfo{pages}{5059--5069}. \DOIprefix\doi{10.18653/v1/P19-1499}.
\bibitem[{Yuan et~al.(2020)Yuan, Wang, and Li}]{yuan2020sesg}
\bibinfo{author}{R.~Yuan}, \bibinfo{author}{Z.~Wang}, \bibinfo{author}{W.~Li},
\newblock \bibinfo{title}{{Fact-level Extractive Summarization with
  Hierarchical Graph Mask on BERT}},
\newblock in: \bibinfo{booktitle}{COLING}, \bibinfo{year}{2020}, pp.
  \bibinfo{pages}{5629--5639}.
  \DOIprefix\doi{10.18653/v1/2020.coling-main.493}.
\bibitem[{Huang and Kurohashi(2021)}]{huang2021sedg}
\bibinfo{author}{Y.~J. Huang}, \bibinfo{author}{S.~Kurohashi},
\newblock \bibinfo{title}{{Extractive Summarization Considering Discourse and
  Coreference Relations based on Heterogeneous Graph}},
\newblock in: \bibinfo{booktitle}{EACL}, \bibinfo{year}{2021}, pp.
  \bibinfo{pages}{3046--3052}. \DOIprefix\doi{10.18653/v1/2021.eacl-main.265}.
\bibitem[{Cui et~al.(2020)Cui, Hu, and Liu}]{cui2020seug}
\bibinfo{author}{P.~Cui}, \bibinfo{author}{L.~Hu}, \bibinfo{author}{Y.~Liu},
\newblock \bibinfo{title}{{Enhancing Extractive Text Summarization with
  Topic-Aware Graph Neural Networks}},
\newblock in: \bibinfo{booktitle}{COLING}, \bibinfo{year}{2020}, pp.
  \bibinfo{pages}{5360--5371}.
  \DOIprefix\doi{10.18653/v1/2020.coling-main.468}.
\bibitem[{Xu et~al.(2020)Xu, Gan, Cheng, and Liu}]{xu2020sedg}
\bibinfo{author}{J.~Xu}, \bibinfo{author}{Z.~Gan}, \bibinfo{author}{Y.~Cheng},
  \bibinfo{author}{J.~Liu},
\newblock \bibinfo{title}{{Discourse-Aware Neural Extractive Text
  Summarization}},
\newblock in: \bibinfo{booktitle}{ACL}, \bibinfo{year}{2020}, pp.
  \bibinfo{pages}{5021--5031}. \DOIprefix\doi{10.18653/v1/2020.acl-main.451}.
\bibitem[{Zheng et~al.(2019)Zheng, Sun, Li, and Muthuswamy}]{zheng2019meng}
\bibinfo{author}{X.~Zheng}, \bibinfo{author}{A.~Sun}, \bibinfo{author}{J.~Li},
  \bibinfo{author}{K.~Muthuswamy},
\newblock \bibinfo{title}{{Subtopic-driven Multi-Document Summarization}},
\newblock in: \bibinfo{booktitle}{EMNLP-IJCNLP}, \bibinfo{year}{2019}, pp.
  \bibinfo{pages}{3153--3162}. \DOIprefix\doi{10.18653/v1/D19-1311}.
\bibitem[{Wang et~al.(2020)Wang, Liu, Zheng, Qiu, and Huang}]{wang2020mesg}
\bibinfo{author}{D.~Wang}, \bibinfo{author}{P.~Liu},
  \bibinfo{author}{Y.~Zheng}, \bibinfo{author}{X.~Qiu},
  \bibinfo{author}{X.~Huang},
\newblock \bibinfo{title}{{Heterogeneous Graph Neural Networks for Extractive
  Document Summarization}},
\newblock in: \bibinfo{booktitle}{ACL}, \bibinfo{year}{2020}, pp.
  \bibinfo{pages}{6209--6219}. \DOIprefix\doi{10.18653/v1/2020.acl-main.553}.
\bibitem[{Erkan and Radev(2004)}]{erkan2004lexrank}
\bibinfo{author}{G.~Erkan}, \bibinfo{author}{D.~R. Radev},
\newblock \bibinfo{title}{{LexRank: Graph-Based Lexical Centrality as Salience
  in Text Summarization}},
\newblock \bibinfo{journal}{Journal of Artificial Intelligence Research}
  \bibinfo{volume}{22} (\bibinfo{year}{2004}) \bibinfo{pages}{457--479}.
  \DOIprefix\doi{10.48550/arXiv.1109.2128}.
\bibitem[{Li et~al.(2020)Li, Xiao, Liu, Wu, Wang, and Du}]{li2020masdg}
\bibinfo{author}{W.~Li}, \bibinfo{author}{X.~Xiao}, \bibinfo{author}{J.~Liu},
  \bibinfo{author}{H.~Wu}, \bibinfo{author}{H.~Wang}, \bibinfo{author}{J.~Du},
\newblock \bibinfo{title}{{Leveraging Graph to Improve Abstractive
  Multi-Document Summarization}},
\newblock in: \bibinfo{booktitle}{ACL}, \bibinfo{year}{2020}, pp.
  \bibinfo{pages}{6232--6243}. \DOIprefix\doi{10.18653/v1/2020.acl-main.555}.
\bibitem[{Flanigan et~al.(2014)Flanigan, Thomson, Carbonell, Dyer, and
  Smith}]{flanigan2014jamr}
\bibinfo{author}{J.~Flanigan}, \bibinfo{author}{S.~Thomson},
  \bibinfo{author}{J.~Carbonell}, \bibinfo{author}{C.~Dyer},
  \bibinfo{author}{N.~A. Smith},
\newblock \bibinfo{title}{{A Discriminative Graph-Based Parser for the Abstract
  Meaning Representation}},
\newblock in: \bibinfo{booktitle}{ACL}, \bibinfo{year}{2014}, pp.
  \bibinfo{pages}{1426--1436}. \DOIprefix\doi{10.3115/v1/P14-1134}.
\bibitem[{Hermann et~al.(2015)Hermann, Ko\v{c}isk\'{y}, Grefenstette, Espeholt,
  Kay, Suleyman, and Blunsom}]{hermann2015conlppar}
\bibinfo{author}{K.~M. Hermann}, \bibinfo{author}{T.~Ko\v{c}isk\'{y}},
  \bibinfo{author}{E.~Grefenstette}, \bibinfo{author}{L.~Espeholt},
  \bibinfo{author}{W.~Kay}, \bibinfo{author}{M.~Suleyman},
  \bibinfo{author}{P.~Blunsom},
\newblock \bibinfo{title}{{Teaching Machines to Read and Comprehend}},
\newblock in: \bibinfo{booktitle}{NeurIPS}, \bibinfo{year}{2015}, pp.
  \bibinfo{pages}{1693--1701}. \DOIprefix\doi{10.48550/arXiv.1506.03340}.
\bibitem[{Dozat and Manning(2017)}]{dozat2017neudpar}
\bibinfo{author}{T.~Dozat}, \bibinfo{author}{C.~D. Manning},
\newblock \bibinfo{title}{{Deep Biaffine Attention for Neural Dependency
  Parsing}},
\newblock in: \bibinfo{booktitle}{ICLR}, \bibinfo{year}{2017}, pp.
  \bibinfo{pages}{1--8}. \DOIprefix\doi{10.48550/arXiv.1611.01734}.
\bibitem[{Ji and Eisenstein(2014)}]{ji2014discpar}
\bibinfo{author}{Y.~Ji}, \bibinfo{author}{J.~Eisenstein},
\newblock \bibinfo{title}{{Representation Learning for Text-level Discourse
  Parsing}},
\newblock in: \bibinfo{booktitle}{ACL}, \bibinfo{year}{2014}, pp.
  \bibinfo{pages}{13--24}. \DOIprefix\doi{10.3115/v1/P14-1002}.
\bibitem[{Miller(1995)}]{miller1995wordnet}
\bibinfo{author}{G.~A. Miller},
\newblock \bibinfo{title}{{WordNet: A Lexical Database for English}},
\newblock \bibinfo{journal}{Communications of the ACM} \bibinfo{volume}{38}
  (\bibinfo{year}{1995}) \bibinfo{pages}{39--41}.
  \DOIprefix\doi{10.1145/219717.219748}.
\bibitem[{Ruppenhofer et~al.(2006)Ruppenhofer, Ellsworth, Petruck, Johnson, and
  Scheffczyk}]{ruppenhofer2006framenet}
\bibinfo{author}{J.~Ruppenhofer}, \bibinfo{author}{M.~Ellsworth},
  \bibinfo{author}{M.~R.~L. Petruck}, \bibinfo{author}{C.~R. Johnson},
  \bibinfo{author}{J.~Scheffczyk},
\newblock \bibinfo{title}{{FrameNet II}: Extended theory and practice},
\newblock \bibinfo{journal}{FrameNet Project}  (\bibinfo{year}{2006}).
\bibitem[{Speer and Havasi(2012)}]{speerhavasi2012conceptnet}
\bibinfo{author}{R.~Speer}, \bibinfo{author}{C.~Havasi},
\newblock \bibinfo{title}{{Representing General Relational Knowledge in
  ConceptNet 5}},
\newblock in: \bibinfo{booktitle}{LREC}, \bibinfo{year}{2012}, pp.
  \bibinfo{pages}{3679--3686}.
\bibitem[{Vrande\v{c}i\'{c} and Kr\"{o}tzsch(2014)}]{vrandei2014wikidata}
\bibinfo{author}{D.~Vrande\v{c}i\'{c}}, \bibinfo{author}{M.~Kr\"{o}tzsch},
\newblock \bibinfo{title}{{Wikidata: A Free Collaborative Knowledgebase}},
\newblock \bibinfo{journal}{Communications of the ACM} \bibinfo{volume}{57}
  (\bibinfo{year}{2014}) \bibinfo{pages}{78--85}.
  \DOIprefix\doi{10.1145/2629489}.
\bibitem[{Angeli et~al.(2015)Angeli, Premkumar, and
  Manning}]{angeli2015lsopenie}
\bibinfo{author}{G.~Angeli}, \bibinfo{author}{M.~J.~J. Premkumar},
  \bibinfo{author}{C.~D. Manning},
\newblock \bibinfo{title}{{Leveraging Linguistic Structure For Open Domain
  Information Extraction}},
\newblock in: \bibinfo{booktitle}{ACL-IJCNLP}, \bibinfo{year}{2015}, pp.
  \bibinfo{pages}{344--354}. \DOIprefix\doi{10.3115/v1/P15-1034}.
\bibitem[{Stanovsky et~al.(2018)Stanovsky, Michael, Zettlemoyer, and
  Dagan}]{stanovsky2018sopenie}
\bibinfo{author}{G.~Stanovsky}, \bibinfo{author}{J.~Michael},
  \bibinfo{author}{L.~Zettlemoyer}, \bibinfo{author}{I.~Dagan},
\newblock \bibinfo{title}{{Supervised Open Information Extraction}},
\newblock in: \bibinfo{booktitle}{NAACL-HLT}, \bibinfo{year}{2018}, pp.
  \bibinfo{pages}{885--895}. \DOIprefix\doi{10.18653/v1/N18-1081}.
\bibitem[{Blei et~al.(2003)Blei, Ng, and Jordan}]{blei2003lda}
\bibinfo{author}{D.~M. Blei}, \bibinfo{author}{A.~Y. Ng},
  \bibinfo{author}{M.~I. Jordan},
\newblock \bibinfo{title}{{Latent Dirichlet Allocation}},
\newblock \bibinfo{journal}{Journal of Machine Learning Research}
  \bibinfo{volume}{3} (\bibinfo{year}{2003}) \bibinfo{pages}{993--1022}.
\bibitem[{Miao et~al.(2017)Miao, Grefenstette, and Blunsom}]{miao2017ntm}
\bibinfo{author}{Y.~Miao}, \bibinfo{author}{E.~Grefenstette},
  \bibinfo{author}{P.~Blunsom},
\newblock \bibinfo{title}{{Discovering Discrete Latent Topics with Neural
  Variational Inference}},
\newblock in: \bibinfo{booktitle}{ICML}, volume~\bibinfo{volume}{70},
  \bibinfo{year}{2017}, pp. \bibinfo{pages}{2410--2419}.
  \DOIprefix\doi{10.48550/arXiv.1706.00359}.
\bibitem[{Liu et~al.(2021)Liu, Wan, He, Peng, and Yu}]{liu2021kgbart}
\bibinfo{author}{Y.~Liu}, \bibinfo{author}{Y.~Wan}, \bibinfo{author}{L.~He},
  \bibinfo{author}{H.~Peng}, \bibinfo{author}{P.~S. Yu},
\newblock \bibinfo{title}{{KG-BART: Knowledge Graph-Augmented BART for
  Generative Commonsense Reasoning}},
\newblock in: \bibinfo{booktitle}{AAAI}, volume~\bibinfo{volume}{35},
  \bibinfo{year}{2021}, pp. \bibinfo{pages}{6418--6425}.
  \DOIprefix\doi{10.48550/arXiv.2009.12677}.
\bibitem[{Liu et~al.(2015)Liu, Flanigan, Thomson, Sadeh, and
  Smith}]{liu2015sasg}
\bibinfo{author}{F.~Liu}, \bibinfo{author}{J.~Flanigan},
  \bibinfo{author}{S.~Thomson}, \bibinfo{author}{N.~Sadeh},
  \bibinfo{author}{N.~A. Smith},
\newblock \bibinfo{title}{{Toward Abstractive Summarization Using Semantic
  Representations}},
\newblock in: \bibinfo{booktitle}{NAACL-HLT}, \bibinfo{year}{2015}, pp.
  \bibinfo{pages}{1077--1086}. \DOIprefix\doi{10.3115/v1/N15-1114}.
\bibitem[{Bordes et~al.(2013)Bordes, Usunier, Garcia-Dur\'{a}n, Weston, and
  Yakhnenko}]{bordes2013transe}
\bibinfo{author}{A.~Bordes}, \bibinfo{author}{N.~Usunier},
  \bibinfo{author}{A.~Garcia-Dur\'{a}n}, \bibinfo{author}{J.~Weston},
  \bibinfo{author}{O.~Yakhnenko},
\newblock \bibinfo{title}{{Translating Embeddings for Modeling Multi-Relational
  Data}},
\newblock in: \bibinfo{booktitle}{NeurIPS}, volume~\bibinfo{volume}{26},
  \bibinfo{year}{2013}, pp. \bibinfo{pages}{2787--2795}.
\bibitem[{Abdi et~al.(2017)Abdi, Idris, Alguliyev, and
  Aliguliyev}]{abdi2017mekg}
\bibinfo{author}{A.~Abdi}, \bibinfo{author}{N.~Idris}, \bibinfo{author}{R.~M.
  Alguliyev}, \bibinfo{author}{R.~M. Aliguliyev},
\newblock \bibinfo{title}{{Query-based multi-documents summarization using
  linguistic knowledge and content word expansion}},
\newblock \bibinfo{journal}{Soft Computing} \bibinfo{volume}{21}
  (\bibinfo{year}{2017}) \bibinfo{pages}{1785--1801}.
  \DOIprefix\doi{10.1007/s00500-015-1881-4}.
\bibitem[{Xie et~al.(2021)Xie, Sun, Deng, Li, and Ding}]{xie2021inconerror}
\bibinfo{author}{Y.~Xie}, \bibinfo{author}{F.~Sun}, \bibinfo{author}{Y.~Deng},
  \bibinfo{author}{Y.~Li}, \bibinfo{author}{B.~Ding},
\newblock \bibinfo{title}{{Factual Consistency Evaluation for Text
  Summarization via Counterfactual Estimation}},
\newblock in: \bibinfo{booktitle}{EMNLP}, \bibinfo{year}{2021}, pp.
  \bibinfo{pages}{100--110}.
  \DOIprefix\doi{10.18653/v1/2021.findings-emnlp.10}.
\bibitem[{Jain et~al.(2020)Jain, van Zuylen, Hajishirzi, and
  Beltagy}]{jain2020scirex}
\bibinfo{author}{S.~Jain}, \bibinfo{author}{M.~van Zuylen},
  \bibinfo{author}{H.~Hajishirzi}, \bibinfo{author}{I.~Beltagy},
\newblock \bibinfo{title}{{SciREX: A Challenge Dataset for Document-Level
  Information Extraction}},
\newblock in: \bibinfo{booktitle}{ACL}, \bibinfo{year}{2020}, pp.
  \bibinfo{pages}{7506--7516}. \DOIprefix\doi{10.18653/v1/2020.acl-main.670}.
\bibitem[{Wu et~al.(2020)Wu, Koncel-Kedziorski, Ostendorf, and
  Hajishirzi}]{wu2020kgie}
\bibinfo{author}{Z.~Wu}, \bibinfo{author}{R.~Koncel-Kedziorski},
  \bibinfo{author}{M.~Ostendorf}, \bibinfo{author}{H.~Hajishirzi},
\newblock \bibinfo{title}{{Extracting Summary Knowledge Graphs from Long
  Documents}},
\newblock \bibinfo{journal}{arXiv} \bibinfo{volume}{abs/2009.09162}
  (\bibinfo{year}{2020}) \bibinfo{pages}{1--8}.
  \DOIprefix\doi{10.48550/arXiv.2009.09162}.
\bibitem[{Pai and Costabello(2021)}]{pai2021focuse}
\bibinfo{author}{S.~Pai}, \bibinfo{author}{L.~Costabello},
\newblock \bibinfo{title}{{Learning Embeddings from Knowledge Graphs With
  Numeric Edge Attributes}},
\newblock in: \bibinfo{booktitle}{IJCAI}, \bibinfo{year}{2021}, pp.
  \bibinfo{pages}{2869--2875}. \DOIprefix\doi{10.24963/ijcai.2021/395}.
\bibitem[{Sharma et~al.(2019)Sharma, Huang, Hu, and Wang}]{sharma2019sang}
\bibinfo{author}{E.~Sharma}, \bibinfo{author}{L.~Huang},
  \bibinfo{author}{Z.~Hu}, \bibinfo{author}{L.~Wang},
\newblock \bibinfo{title}{{An Entity-Driven Framework for Abstractive
  Summarization}},
\newblock in: \bibinfo{booktitle}{EMNLP-IJCNLP}, \bibinfo{year}{2019}, pp.
  \bibinfo{pages}{3280--3291}. \DOIprefix\doi{10.18653/v1/D19-1323}.
\bibitem[{Liu et~al.(2021)Liu, Yuan, Fu, Jiang, Hayashi, and
  Neubig}]{liu2021pre}
\bibinfo{author}{P.~Liu}, \bibinfo{author}{W.~Yuan}, \bibinfo{author}{J.~Fu},
  \bibinfo{author}{Z.~Jiang}, \bibinfo{author}{H.~Hayashi},
  \bibinfo{author}{G.~Neubig},
\newblock \bibinfo{title}{{Pre-train, Prompt, and Predict: A Systematic Survey
  of Prompting Methods in Natural Language Processing}},
\newblock \bibinfo{journal}{arXiv preprint arXiv:2107.13586}
  (\bibinfo{year}{2021}).

\end{thebibliography}

\end{document}